# Super-Resolution and Image Re-projection for Iris Recognition

Eduardo Ribeiro
Federal University of Tocantins
Palmas, Brazil
uft.eduardo@uft.edu.br

Andreas Uhl
niversity of Salzburg
Salzburg, Austria
uhl@cosy.sbg.ac.at

Fernando Alonso-Fernandez
Halmstad University
Halmstad, Sweden
feralo@hh.se

## Abstract

*Several recent works have addressed the ability of deep learning to disclose rich, hierarchical and discriminative models for the most diverse purposes. Specifically in the super-resolution field, Convolutional Neural Networks (CNNs) using different deep learning approaches attempt to recover realistic texture and fine grained details from low resolution images. In this work we explore the viability of these approaches for iris Super-Resolution (SR) in an iris recognition environment. For this, we test different architectures with and without a so called image re-projection to reduce artifacts applying it to different iris databases to verify the viability of the different CNNs for iris super-resolution. Results show that CNNs and image re-projection can improve the results specially for the accuracy of recognition systems using a complete different training database performing the transfer learning successfully.*

## 1. Introduction

Iris recognition is currently considered one of the most reliable techniques for human identification, mainly due its stability and high degree of freedom in texture [1]. One of the research community concerns is to provide systems that allows iris capture in a more relaxed way, such as through surveillance cameras or mobile devices allowing long acquisition distance between the iris and the sensor, in addition to allowing less powerful sensors [25].

In iris recognition systems, image resolution plays a key role to represent all the necessary information for recognition, which means that the lower the resolution, the greater the degradation of recognition performance. However a high pixel density does not automatically lead to a good image quality. Generally a conventional image up-sampling will only replicate or interpolate the information already contained in the image resulting in a large, blurred image of poor quality [20].

The main goal of super-resolution is to recover from one (single image) or more Low Resolution (LR) images a High Resolution (HR) image that has more pixels than the original image producing a more detailed and realistic image while being faithful to the low resolution image(s) [4]. The LR image $\bar{Y}$ is modeled as the corresponding HR $\bar{X}$ image manipulated by blurring ($B$), warping ($W$) and down-sampling ($D$) as $\bar{X} = DBW\bar{Y} + \bar{n}$ ($n$ represents additive noise). For simplicity, some works omit the warp matrix and noise, leading to $\bar{X} = DB\bar{Y}$ [1].

In this work, we perform a single-image super-resolution via CNNs that will do, during the training phase, a map between the HR training images with high frequency information and the LR images with low-frequency information. It is expected that this mapping will produce more detailed HR images with more texture in the iris patterns.

Many studies have been performed in the super-resolution area in recent years. Some surveys about super-resolution can be found in [13] [24]. Recently with the popularization of Convolutional Neural Networks, several methods for super-resolution have been proposed obtaining excellent results in relation to the quality of the images as for example the sparse representation through CNNs proposed by Wang et. al. [22] or the denoising CNN known as SRCNN proposed by Dong et. al. [9]. On this basis, several others CNNs have been proposed including the Very Deep CNN (VDCNN) proposed by Jiwon Kim et. al [10] inspired by the VGG-net [19], the Generative Adversarial Networks (SRGAN) proposed by Ledig et. al. [11] and the Fast and Accurate Image super-resolution by Deep CNN with Skip Connection (DCSCN) proposed by Yamanaka et. al [23]. These three methods will be used in this work to perform iris super-resolution and will be better explained in the next sections.

Despite the great interest of Super Resolution and the use of deep learning in Biometrics [7] [14], the use of SR for iris recognition is still raising many questions, mainly because SR is focused on the generation of high quality images while iris recognition focuses on the best recognition performance itself [15].

In some previous works [17] [18] were evaluated different CNNs configurations with different training databases, showing that it is possible to apply CNNs for recognition of iris images in low resolution. Specifically in this work, we will further explore techniques already shown to be useful for natural images (VDCNNs and SRGANs) and add another recently proposed method (DCSCNs) to verify if the good performance of these methods in natural images in terms of photo-realism is also valid for iris images in the iris recognition context. Additionally we will test if the image re-projection proposed in [1] to reduce artifacts can also improve the results that it was not tested in the previous works.

## 2. Methodology

The main focus of deep learning image super resolution is the mapping between the LR images and the HR images during CNN training. In this case, a training database is chosen and the HR images are prepared. In this process the images are downscaled to one or more factors followed by a upscaling using bicubic interpolation to the same size as the HR image. During training, the LR image or patch is fed into the CNN to obtain a reconstructed image that has to be as similar as possible to the HR image, in this case, the ground truth. In this off-line phase, the mapping between the high-resolution (HR) image and the low-resolution (LR) image is made using the chosen CNN architecture with the chosen database as shown in the Figure 1 (off-line phase). In figure 1, to achieve the factor 2, it is shown that the image will be interpolate and pass through the trained CNN just one time. To achieve greater factors, the images have to be re-inserted in the CNN $log_2(n) times$, where n is the desired factor.

The weight adjustment is done using a back-propagation algorithm that will depend on the architecture and of the loss function chosen. After training, the CNN will be applied to a target database that is, in the case of this work, an iris database also called target database. It can be said that the propagation of the LR image through the CNN is a pre-processing step before the iris recognition, which means that the HR images to be used in the iris recognition system are the reconstruction of LR resolution iris images trough the CNN as is shown in Figure 1 (on-line phase).

### 2.1. Training and Target/Testing Databases

The training database chosen for this work is the Describable Texture Dataset (**DTD**) with 5640 images of sizes range between $300 \times 300$ and $640 \times 640$ categorized in 47 classes [6]. This database was tested in some previous works ([17] [18]) and among several other tested databases, including natural image databases, texture databases and iris databases, DTD lead to the best results for very low resolution images.

Moreover, in this work we use two target databases, the first one is a popular database widely used on biometrics: the CASIA Interval V3 database. The second database was chosen to explore the use of CNN's in a real world situation where the images are captured from mobile devices and is called Visible Spectrum Smart-phone Iris (VSSIRIS [16]) database.

The CASIA Interval V3 database contains 2655 NIR images of $280 \times 320$ pixels from 249 subjects captured in two sessions with a self-developed close-up camera, resulting in 396 different eyes which will be considered as different subjects for this work. As a pre-processing step, all images are resized using bicubic interpolation in order to have the same sclera radius. After this, a cropping around the pupil in a square region of size 231x231 is performed and the images that do not fit into this cropping (e.g. if the iris is close to a margin) are discarded. We divide the remaining 1872 images into two sets: the first 116 users (925 images) are used for the training and the remaining 133 users (947 images) are used as the target database. Although CNNs do not use this iris training set we decided to keep this procedure in order to compare the methods with other methods that use this framework.

The Visible Spectrum Smart-phone Iris (VSSIRIS) database contains images obtained by using two different mobile phone devices: Apple Iphone 5S ($3264x2448$ pixels) and Nokia Lumia 1020 ($3072x1728$ pixels). Five images of the two eyes from 28 subjects were acquired from each device totalizing 560 images. Also, in this database all the images are resized to have the same sclera radius followed by a cropping around the pupil in a square region of size 319x319 pixels as a pre-processing step. The downsampling factor used for this experiment is the factor 1/22 following the previous studies in [21] and [1] to generate a very small iris region (13x13 pixels) for the real-world low resolution simulation.

### 2.2. CNN Architectures

For this work we test three different CNN architectures designed specifically for Singe-Image Super-Resolution: Very Deep Super-Resolution CNN (**VDCNN**), Super Resolution Adversarial CNN (**SRGAN**) and the Super-Resolution Deep CNN with Skip Connection and Network in Network (DCSCN).

- The VDCNN is proposed in [10] is based on the VGG-net containing 20 layers with the same parameterization (except for the first and the last layers): 64 filters of size $3 \times 3 \times 64$. The loss function used in the training is the Mean Squared Error between the residual input error and the residual ground truth.

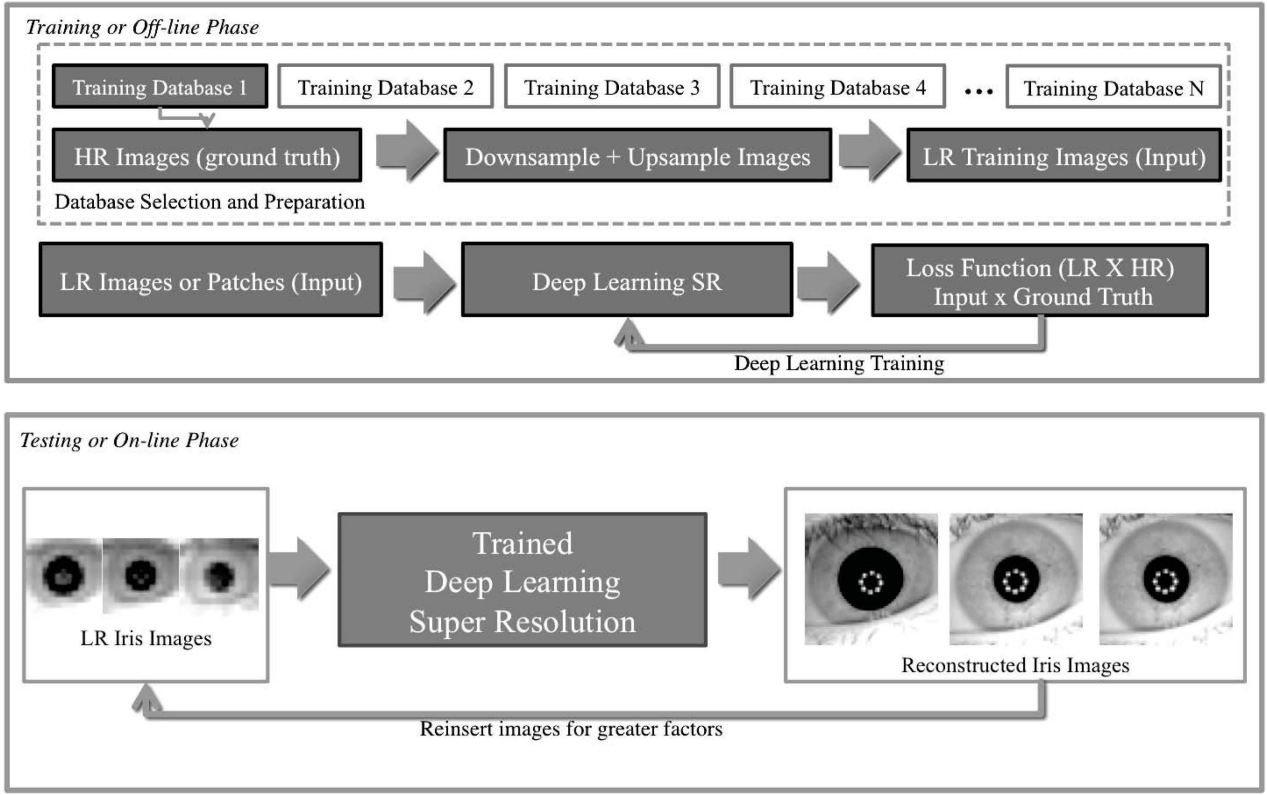

Figure 1: General overview of the training and reconstruction method for Iris Super Resolution using CNNs proposed for this work.

- The SRGAN approach proposed in [11] relies on two different CNNs (generator and discriminator architectures) where the generator network is formed by a series of residual blocks with identical layout: two convolutional layers of size with $3 \times 3 \times 64$ followed by a batch-normalization and ParametricRELU layers as activation function while the discriminator network is also based on the VGG-net as the VDCNN.

- The DCSCN approach proposed in [23] uses a Deep CNN in combination with a Residual net optimizing the number of layers and filters. Besides that, this method uses a scheme called Skip Connection Layers in order to decrease the computation effort using these layers as feature extractors for image features on both local and global areas.

### 2.3. Image Re-Projection

In this work we test a post-processing step to the CNN result called image re-projection following the work of [5] applied to the reconstructed image $\bar{Y}'$ trying to reduce artifacts and making the post-processed image more similar to the HR image $\bar{X}'$ as possible. In this phase, the image $\bar{Y}'$ is re-projected to $\bar{X}$ via $\bar{Y}^{T+1} = \bar{Y}^T - \tau U(B(DB\bar{Y}^T - \bar{X}))$ where $U$ is the up-sampling matrix [1]. The process stops when $|\bar{Y}^{T+1} - \bar{Y}^T|$ is smaller than a threshold ($10^{-5}$ in our experiments). We also use $\tau = 0.02$.

### 2.4. Experimental Framework

For the experiments we consider two criteria for evaluation: the quality of reconstructed images and the accuracy of iris recognition. With these metrics we are able to verify if the better quality of the reconstructed images can directly influence recognition performance.

The quality assessment algorithms used are: the Peak Signal to Noise Ratio (PSNR), the Structural Similarity Index Measure (SSIM) and the Feature Similarity Index for Image Quality Assessment (FSIM) from [26] where a high score reflects a high quality using the HR image as the reference image.

For iris recognition we use two different systems based on 1D log-Gabor filters (LG) [12], and the SIFT operator (SIFT) [3]. In the LG method, first the iris is unwrapped to a normalized rectangle using the Daugman's rubber sheet model [8] then a 1D Log-Gabor wavelet is applied with phase binary quantization to 4 levels. The matching is done using the normalized Hamming distance. In the SIFT method, SIFT keypoints are extracted from the iris region without unwrapping and the matching is done by counting the number of matched keypoints between two images. We then carry out fusion experiments using linear logistic regression. Given $N$ comparators ($N=2$ in our case) which output the scores ($s_{1j}, s_{2j}, ...s_{Nj}$) for an input trial $j$, a linear fusion is: $f_j = a_0 + a_1 \cdot s_{1j} + a_2 \cdot s_{2j} + ... + a_N \cdot s_{Nj}$. The weights $a_0, a_1, ...a_N$ are trained via logistic regression as described in [2]. We use this trained fusion approach because it has shown better performance than simple fusion rules (like the mean or the sum rule) in previous works.

In the case of this work, in the target databases, each eye is considered as a different user. Genuine matches are obtained by comparing each image of a user to the remaining images of the same user, avoiding symmetric matches. Impostor matches are obtained by comparing the first image of a user to the second image of the remaining users. With this procedure, we obtain 2,607 genuine and 19,537 impostor scores for the the CASIA Interval V3 database. For the Visible Spectrum Smart-phone Iris database using the same procedure we obtain 560 genuine and 3,018 impostor scores per smartphone.

We compare our results with two classic approaches: bilinear and bicubic interpolation and a method used in [1] for iris-super resolution called PCA hallucination of local patches based on the algorithm for face images of [5] where a PCA eigen-transformation is conducted in the set of LR basis patches to use the weights provided by the projection of the eigen-patches to reconstruct the images.

## 3. Experimental Results

In both test sets (CASIA Interval V3 database and VS-SIRIS database) the original images in High Resolution (HR) were used as reference images. Then the images were downsampled via bicubic interpolation for the factor 1/2,1/4, 1/8 and 1/16 for the CASIA Interval V3 and 1/18 for the VSSIRIS database. This approach using simulated LR images follows previous works in super resolution research mainly because there is still a lack of real world low resolution databases with high resolution reference images. In both databases, for the image quality tests we also extract the normalized iris region to analyse the results in the region of interest.

Table 1 presents the quality assessment results for the CASIA Interval V3 database. It can be seen that, for the full image the Deep Learning approaches surpasses the results from the classical algorithms and PCA for almost all factors. Analysing just the iris region (that is the most important region for our experiment), it can be seen that the deep learning methods also are superior to the compared methods especially for the VDCNN. Also, in both scenarios the image re-projection improves the results. Table 2 presents the quality assessment results for the VSSIRIS database. In general the results are better than the classical algorithms for the full image and the iris region. It can be noticed that in terms of quality the results from the PCA approach are similar to the VDCNN. The advantage of the deep learning approach in detriment to the PCA method in that the CNN is trained with a complete different database (no iris or infrared images) while PCA framework uses images from the same database for training the algorithm.

For the recognition experiment we consider a totally uncontrolled scenario where enrolment and query samples are taken from the reconstructed super-resolution results (for example, when the user is registered using a cellphone and makes use of the system also using a cellphone camera with certain distance).

Table 3 presents the recognition results for the CASIA iris V3 database. It can be observed that all deep learning approaches present better results than the classical algorithms specially for the SIFT method for which the DCSCN approach presents the best result. It also can be noticed that, similarly to the quality results, the image re-projection also improves the results for recognition in almost all the cases. Besides that, it is interesting to notice that the best quality results among the CNNs reconstruction does not necessarily lead to the best recognition performance and the reconstructed results surpasses the results from the original images for the factor 1/2 and 1/4.

Table 4 presents the recognition performance for the VSSIRIS database. Using the Iphone images it can be noticed that differently to the first database, the best result using LG is from a Deep Learning approach (DCSNC) which does not benefit from the image re-projection. Using the other two recognition approaches, the best result is from the PCA method. For the Nokia images the best results came from the classical approaches for the LG and SIFT algorithms. When we make the fusion with the features the DCSCN method presents better results.

## 4. Conclusions and Future Work

The use of mobile devices and more relaxed acquisition condition for iris recognition are becoming increasingly common. In this paper, we test different CNN architectures and image re-projection to increase the resolution of infraded (NIR) and mobile phone iris images. We compare the results using quality assessment algorithms and recognition methods with two classical approaches (bicubic and bilinear interpolation) as well as with the PCA approach. The experiments show, that the CNN reconstruction methods improve the results especially in the recognition context for different scaling factors including very low resolution images. Considering the very high reduction factor, it can be seen that in some cases, the EER was satisfactory which is very important for systems that work under very low stor-age or transmission capabilities.

| | | Bilinear | Bicubic | PCA | Without Re-projection ||| With Re-projection |||
| | | | | | DCGAN | SRGAN | VDCNN | DCSN | SRGAN | VDCNN |
|---|---|---|---|---|---|---|---|---|---|---|
| Scaling | | | | | Full image |||||||
| 115×115 1/2 | psnr | 33.00 | 34.04 | 34.65 | 35.09 | 23.31 | 36.10 | 36.25 | 35.50 | **36.75** |
| | ssim | 0.91 | 0.93 | 0.93 | 0.94 | 0.90 | 0.95 | 0.95 | 0.94 | **0.96** |
| 57×57 1/4 | psnr | 28.36 | 29.18 | 29.90 | 30.19 | 23.57 | 29.55 | 30.91 | 29.53 | **31.02** |
| | ssim | 0.79 | 0.80 | 0.81 | 0.82 | 0.77 | 0.82 | **0.84** | 0.81 | **0.84** |
| 29×29 1/8 | psnr | 24.86 | 25.33 | **26.73** | 25.95 | 21.83 | 26.23 | 26.13 | 25.44 | 26.43 |
| | ssim | 0.69 | 0.70 | 0.71 | 0.71 | 0.65 | 0.72 | 0.72 | 0.67 | **0.73** |
| 15×15 1/16 | psnr | 22.39 | 22.86 | **24.32** | 23.45 | 21.24 | 23.81 | 23.62 | 23.31 | 23.88 |
| | ssim | 0.64 | 0.64 | **0.66** | 0.65 | 0.60 | 0.66 | 0.65 | 0.62 | **0.66** |
| Scaling | | | | | Iris region |||||||
| 115×115 1/2 | psnr | 36.94 | 38.22 | 39.15 | 39.42 | 22.88 | 40.55 | 41.33 | 40.43 | **41.79** |
| | ssim | 0.96 | 0.97 | 0.97 | 0.97 | 0.94 | **0.98** | **0.98** | **0.98** | **0.98** |
| 57×57 1/4 | psnr | 31.64 | 32.35 | 32.72 | 33.11 | 24.37 | 32.46 | 33.91 | 32.83 | **34.05** |
| | ssim | 0.85 | 0.87 | 0.88 | 0.89 | 0.85 | 0.88 | **0.91** | 0.89 | **0.91** |
| 29×29 1/8 | psnr | 28.18 | 28.74 | 29.57 | 29.59 | 22.70 | 29.66 | 29.74 | 29.31 | **30.01** |
| | ssim | 0.74 | 0.75 | 0.77 | 0.77 | 0.70 | **0.78** | **0.78** | 0.75 | **0.78** |
| 15×15 1/16 | psnr | 25.16 | 25.64 | 26.99 | 26.47 | 23.27 | 26.96 | 26.61 | **27.17** | 27.12 |
| | ssim | 0.63 | 0.64 | 0.66 | 0.65 | 0.62 | **0.67** | 0.65 | 0.65 | **0.67** |

Table 1: Results of quality assessment algorithms using different CNN architectures comparing to the Bilinear, Bicubic and PCA approach for the CASIA Interval V3 Database.

| | bilinear | bicubic | PCA | Without Re-Projection ||| With Re-Projection |||
| | | | | DCSCN | SRGAN | VDCNN | DCSCN | SRGAN | VDCNN |
|---|---|---|---|---|---|---|---|---|---|
| | | | | Full image |||||||
| psnr | 24.44 | 24.97 | **26.00** | 25.51 | 17.16 | 25.27 | 25.66 | 24.60 | 25.80 |
| ssim | 0.72 | 0.72 | **0.73** | **0.73** | 0.59 | **0.73** | **0.73** | 0.64 | **0.73** |
| | | | | Iris region |||||||
| psnr | 24.35 | 24.89 | **25.45** | 25.15 | 18.08 | 24.90 | 25.30 | 24.67 | **25.45** |
| ssim | 0.62 | 0.64 | **0.67** | 0.65 | 0.54 | 0.65 | 0.66 | 0.61 | **0.67** |

Table 2: Results of quality assessment algorithms using different CNN architectures comparing to the Bilinear, Bicubic and PCA approach for the VSSIRIS Database. The scaling for all images tested for this database is 1/18 or 13X13 pixels.

## Acknowledgment

This project has received funding from the European Union's Horizon 2020 research and innovation program under grant agreement No 700259. This research was partially supported by CNPq-Brazil for Eduardo Ribeiro under grant No. 00736/2014-0.

## References

[1] F. Alonso-Fernandez, R. A. Farrugia, and J. Bigun. Learning-based local-patch resolution reconstruction of iris smartphone images. In *IEEE/IAPR International Joint Conference on Biometrics, IJCB*, October 2017.

|  |  |  |  |  | Without Re-projection | | | With Re-projection | | |
|---|---|---|---|---|---|---|---|---|---|---|
| scaling | Comparator | Bilinear | Bicubic | PCA | DCGAN | SRGAN | VDCNN | DCSCN | SRGAN | VDCNN |
| 115×115 1/2 | LG | **0.61%** | 0.73% | 0.72% | 0.73% | 0.80% | 0.73% | 0.73% | 0.73% | 0.72% |
| | SIFT | **3.01%** | 3.13% | 3.71% | 3.33% | 5.66% | 3.67% | 4.00% | 3.75% | 3.82% |
| | LG+SIFT | **0.61%** | **0.61%** | 0.65% | 0.65% | 0.73% | 0.69% | 0.65% | 0.65% | 0.65% |
| 57×57 1/4 | LG | 0.76% | 0.65% | 0.68% | **0.56%** | 1.03% | 0.65% | 0.66% | 0.69% | 0.69% |
| | SIFT | 4.26% | 3.08% | 3.37% | 2.68% | 7.75% | 2.41% | 2.49% | 4.09% | **2.29%** |
| | LG+SIFT | 0.76% | 0.70% | **0.65%** | **0.65%** | 1.03% | **0.65%** | 0.69% | 0.69% | **0.65%** |
| 29×29 1/8 | LG | 2.38% | 1.88% | 1.18% | 1.38% | 3.05% | 1.03% | 1.45% | 1.41% | *1.00%* |
| | SIFT | 14.82% | 11.60% | 7.54% | 9.89% | 17.01% | 8.24% | 7.73% | 8.63% | **7.06%** |
| | LG+SIFT | 2.30% | 1.73% | 1.00% | 1.19% | 2.85% | 0.92% | 1.20% | 1.38% | *0.90%* |
| 15×15 1/16 | LG | 11.03% | 11.25% | **4.79%** | 7.06% | 7.89% | 5.12% | 10.08% | 6.01% | 5.72% |
| | SIFT | 41.66% | 36.37% | 19.50% | 27.02% | 20.81% | 26.76% | **18.00%** | 21.29% | 21.48% |
| | LG+SIFT | 10.98% | 10.97% | **4.41%** | 6.63% | 7.59% | 5.06% | 8.86% | 5.52% | 5.37% |

Table 3: Verification results (EER) using different CNN architectures comparing to the Bilinear, Bicubic and PCA approaches for the CASIA Interval V3 database. The accuracy result for the original database with no scaling is 0.76% for LG , 4.19% for SIFT and 0.71% for LG+SIFT.

|  | Bilinear | Bicubic | PCA | Without Re-Projection | | | With Re-Projection | | |
|---|---|---|---|---|---|---|---|---|---|
|  |  |  |  | DCSCN | SRGAN | VDCNN | DCSCN | SRGAN | VDCNN |
| iPhone | | | | | | | | | |
| LG | 9.10% | 9.83% | 8.90% | **7.62%** | 8.96% | 7.97% | 8.62% | 9.25% | 8.91% |
| SIFT | 23.55% | 22.80% | **9.30%** | 21.66% | 12.00% | 24.29% | 13.30% | 10.03% | 14.39% |
| LG+SIFT | 8.99% | 9.66% | **4.64%** | 6.43% | 5.97% | 6.77% | 6.44% | 5.71% | 5.71% |
| Nokia | | | | | | | | | |
| LG | **8.03%** | 8.43% | 9.41% | 9.01% | 9.46% | 9.69% | 9.27% | 9.42% | 9.44% |
| SIFT | 26.51% | 29.81% | **11.14%** | 22.47% | 14.10% | 26.62% | 14.70% | 14.10% | 16.16% |
| LG+SIFT | 8.03% | 8.23% | 7.53% | 7.67% | 7.84% | 8.20% | **7.13%** | 8.38% | 8.38% |

Table 4: Verification results (EER) for different methods employed in the VSSIRIS database. The accuracy result for the original database with no scaling is 8.04% (LG), 0.33% (SIFT), 0.03% (LG+SIFT) for iPhone image and 7.47% (LG), 0.68% (SIFT), 0.34% (LG+SIFT) for Nokia images. The scaling for all images tested is 1/18 or 13X13 pixels.

[2] F. Alonso-Fernandez, J. Fierrez, D. Ramos, and J. Gonzalez-Rodriguez. Quality-based conditional processing in multi-biometrics: Application to sensor interoperability. *IEEE Transactions on Systems, Man, and Cybernetics - Part A: Systems and Humans*, 40(6):1168–1179, Nov 2010.

[3] F. Alonso-Fernandez, P. Tome-Gonzalez, V. Ruiz-Albacete, and J. Ortega-Garcia. Iris recognition based on sift features. In *2009 First IEEE International Conference on Biometrics, Identity and Security (BIdS)*, Sept 2009.

[4] S. Baker and T. Kanade. Limits on super-resolution and how to break them. *IEEE Transactions on Pattern Analysis and Machine Intelligence*, 24(9):1167–1183, Sep 2002.

[5] H. Y. Chen and S. Y. Chien. Eigen-patch: Position-patch based face hallucination using eigen transformation. In *2014 IEEE International Conference on Multimedia and Expo (ICME)*, pages 1–6, July 2014.

[6] M. Cimpoi, S. Maji, I. Kokkinos, S. Mohamed, , and A. Vedaldi. Describing textures in the wild. In *Proceedings of the IEEE Conf. on Computer Vision and Pattern Recognition (CVPR)*, 2014.

[7] J. Cui, Y. Wang, J. Huang, T. Tan, and Z. Sun. An iris image synthesis method based on pca and super-resolution. In *Proceedings of the 17th International Conference on Pattern Recognition, 2004. ICPR 2004.*, volume 4, pages 471–474

Vol.4, Aug 2004.

[8] J. Daugman. How iris recognition works. *IEEE Trans. Cir. and Sys. for Video Technol.*, 14(1), Jan. 2004.

[9] C. Dong, C. C. Loy, K. He, and X. Tang. Image super-resolution using deep convolutional networks. *CoRR*, abs/1501.00092, 2015.

[10] J. Kim, J. K. Lee, and K. M. Lee. Accurate image super-resolution using very deep convolutional networks. *CoRR*, abs/1511.04587, 2015.

[11] C. Ledig, L. Theis, F. Huszar, J. Caballero, A. P. Aitken, A. Tejani, J. Totz, Z. Wang, and W. Shi. Photo-realistic single image super-resolution using a generative adversarial network. *CoRR*, abs/1609.04802, 2016.

[12] L. Masek. Recognition of human iris patterns for biometric identification. Technical report, The University of Western Australia, 2003.

[13] K. Nasrollahi and T. B. Moeslund. Super-resolution: a comprehensive survey. *Machine Vision and Applications*, 25(6):1423–1468, Aug 2014.

[14] K. Nguyen, C. Fookes, S. Sridharan, M. Tistarelli, and M. Nixon. Super-resolution for biometrics: A comprehensive survey. *Pattern Recognition*, 78:23 – 42, 2018.

[15] K. Nguyen, S. Sridharan, S. Denman, and C. Fookes. Feature-domain super-resolution framework for gabor-based face and iris recognition. In *2012 IEEE Conference on Computer Vision and Pattern Recognition, Providence, RI, USA, June 16-21, 2012*, pages 2642–2649, 2012.

[16] K. B. Raja, R. Raghavendra, V. K. Vemuri, and C. Busch. Smartphone based visible iris recognition using deep sparse filtering. *Pattern Recognition Letters*, 57(Supplement C):33 – 42, 2015. Mobile Iris CHallenge Evaluation part I (MICHE I).

[17] E. Ribeiro and A. Uhl. Exploring texture transfer learning via convolutional neural networks for iris super resolution. In *Proceedings of the 2017 International Conference of the Biometrics Special Interest Group (BIOSIG'17), Darmstadt, Germany 2017*, LNI. GI / IEEE, 2017.

[18] E. Ribeiro, A. Uhl, F. Alonso-Fernandez, and R. A. Farrugia. Exploring deep learning image super-resolution for iris recognition. In *Proc. of the 25th European Signal Processing Conference (EUSIPCO 2017), Kos Island, Greece, August 28 - September 2, 2017*, 2017.

[19] K. Simonyan and A. Zisserman. Very deep convolutional networks for large-scale image recognition. *CoRR*, abs/1409.1556, 2014.

[20] K. C. P. Tanay, S. Khanna, V. Chandrasekaran, and P. K. Baruah. Fast video super resolution using deep convolutional networks. In *2017 International Conference on Innovations in Information, Embedded and Communication Systems (ICIIECS)*, pages 1–6, March 2017.

[21] N. Wang, D. Tao, X. Gao, X. Li, and J. Li. A comprehensive survey to face hallucination. *International Journal of Computer Vision*, 106(1):9–30, Jan 2014.

[22] Z. Wang, D. Liu, J. Yang, W. Han, and T. S. Huang. Deeply improved sparse coding for image super-resolution. *CoRR*, abs/1507.08905, 2015.

[23] J. Yamanaka, S. Kuwashima, and T. Kurita. Fast and accurate image super resolution by deep cnn with skip connection and network in network. In *Neural Information Processing*, pages 217–225, Cham, 2017. Springer International Publishing.

[24] C.-Y. Yang, C. Ma, and M.-H. Yang. *Single-Image Super-Resolution: A Benchmark*, pages 372–386. Springer International Publishing, Cham, 2014.

[25] X. Zeng, H. Huang, and C. Qi. Expanding training data for facial image super-resolution. *IEEE Transactions on Cybernetics*, 48(2):716–729, Feb 2018.

[26] L. Zhang, L. Zhang, X. Mou, and D. Zhang. Fsim: A feature similarity index for image quality assessment. *IEEE Transactions on Image Processing*, 20(8):2378–2386, Aug 2011.